\documentclass{article} 
\usepackage{natbib}

\usepackage{microtype}
\usepackage{hyperref}
\usepackage{url}
\usepackage{booktabs}
\usepackage{graphicx}

\title{Token-Efficient Leverage Learning in Large Language Models}

\author{Yuanhao Zeng  \\
School of Computer Science\\
Beijing University of Posts and Telecommunications \\
Beijing 100876, P.R. China\\
\texttt{cengyuanhao@bupt.edu.cn} \\
Min Wang \\
College of information science and electronic engineering \\
Zhejiang University \\
Hangzhou 310058, P.R. China \\
\texttt{3210106344@zju.edu.cn} \\
Yihang Wang \\
School of Artificial Intelligence\\
Beijing University of Posts and Telecommunications \\
Beijing 100876, P.R. China \\
\texttt{yihangwang1020@gmail.com} \\
Yingxia Shao \\
School of Computer Science\\
Beijing University of Posts and Telecommunications \\
Beijing 100876, P.R. China\\
\texttt{shaoyx@bupt.edu.cn} \\
}

\begin{document}

\maketitle

\begin{abstract}
Large Language Models (LLMs) have excelled in various tasks but perform better in high-resource scenarios, which presents challenges in low-resource scenarios. Data scarcity and the inherent difficulty of adapting LLMs to specific tasks compound the challenge. To address the twin hurdles, we introduce \textbf{Leverage Learning}. We present a streamlined implement of this methodology called Token-Efficient Leverage Learning (TELL). TELL showcases the potential of Leverage Learning, demonstrating effectiveness across various LLMs and low-resource tasks, ranging from \(10^4\) to \(10^6\) tokens. It reduces task data requirements by up to nearly an order of magnitude compared to conventional Supervised Fine-Tuning (SFT) while delivering competitive performance. With the same amount of task data, TELL leads in improving task performance compared to SFT. We discuss the mechanism of Leverage Learning, suggesting it aligns with quantization hypothesis and explore its promising potential through empirical testing.

\end{abstract}

\section{Introduction}

Large Language Models (LLMs) often struggle with low-resource tasks due to limited representation in pre-training, making it challenging to generalize and adapt.\citep{Aghajanyan2020-fr, Zhang2020-gp, alpaca, Min2022-ei} Supervised Fine-Tuning (SFT) on targeted instruction datasets partially addresses this, but the scarcity of domain-specific data and budget constraints in such settings complicate the process. In-context learning offers a remedy by guiding LLMs through examples within prompts, yet it can be unreliable due to sensitivity to prompt structure and content \citep{Zhang2020-gp, Garcia2023-qq, webson2022promptbased, zhao2021calibrate}.

To better train LLMs for specific tasks, fine-tuning is vital for improving accuracy and aligning with user intentions, but it faces obstacles in low-resource scenarios, including the need for high-quality, domain-specific data and the cost of manual verification to ensure data integrity and mitigate biases \citep{Min2022-ei, Zhou2023-pz, Ho2022-hz, Kim2022-hy}.

Therefore, the challenge we face is to effectively fine-tune LLMs in low-resource environments, where data scarcity and the inherent difficulty of adapting LLMs to specific tasks compound the problem. It is imperative to devise a strategy that can \textbf{leverage} the sparse data at our disposal to facilitate efficient fine-tuning for these low-resource tasks, overcoming the pitfalls of in-context learning, and the complexities of manual data augmentation and verification.

In addressing the fine-tuning challenge for low-resource tasks, we introduce \textbf{Leverage Learning}, a framework inspired by quantization hypothesis \citep{Michaud2023-rs}. This framework is designed to maximize the use of available low-resource task data and encapsulates two key techniques essential for realizing our vision. We present a streamlined version of this approach, called Token-Efficient Leverage Learning (TELL), which showcases the potential of Leverage Learning. TELL employs "anchor prompt" and "extensively shuffle" techniques to fulfill the Leverage Learning. This minimalist approach demonstrates that even with basic techniques, we can significantly diminish the dependence on extensive task-specific data while achieving results comparable to those from fine-tuning with larger datasets.

\begin{figure}[h]
\begin{center}
\includegraphics[width=10cm]{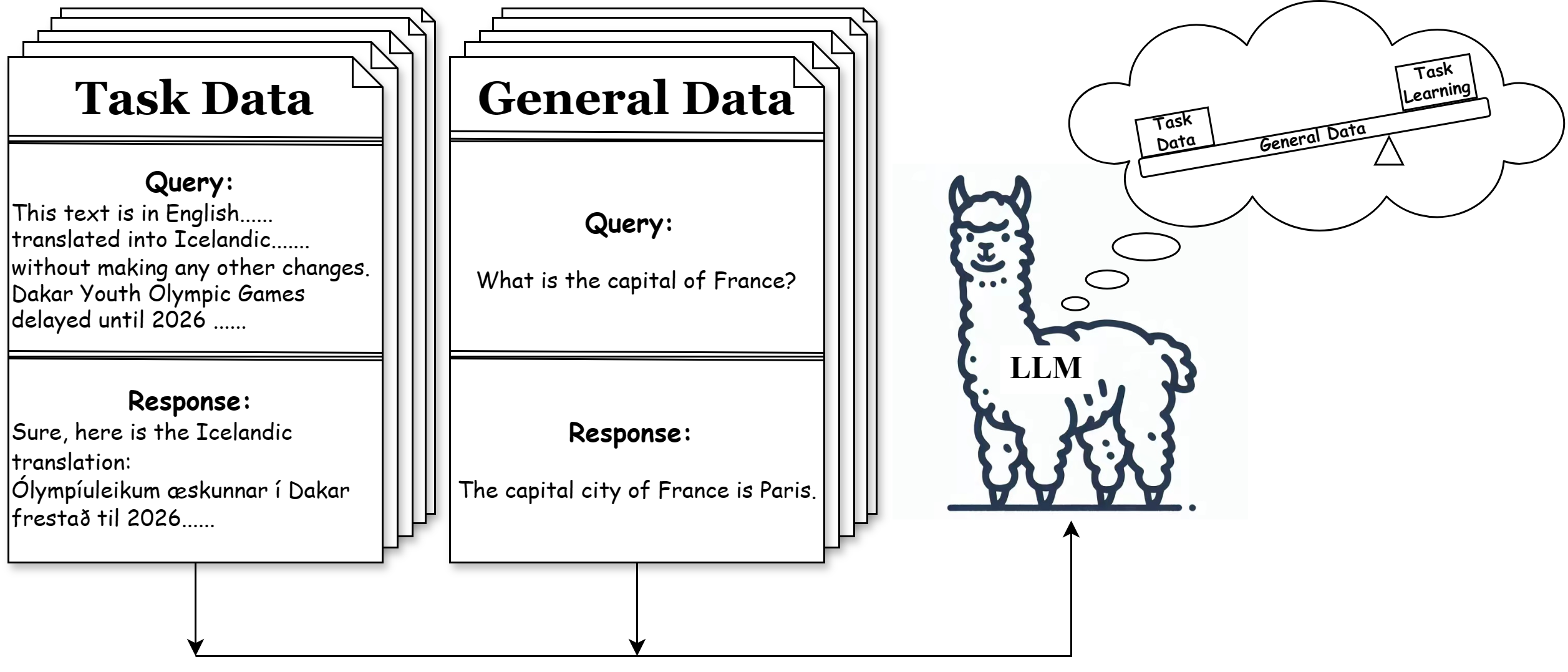}
\end{center}
\caption{An overview of Leverage Learning}
\label{fig:leverageOverview}
\end{figure}

Leverage Learning aims to fully utilize the information in low-resource task data. This methodology minimizes the need for extensive task-specific data by acquiring task-specific capabilities from such data while learning non-specific capabilities from general data. This is akin to a lever, where general data act as the power arm and low-resource data is the input force. This enhances LLMs' efficiency in learning from limited resources, as shown in Figure~\ref{fig:leverageOverview}. To achieve this vision, two issues need solved: First, LLMs must distinguish between general and low-resource tasks to specifically enhance performance on the latter. Second, LLMs need to adopt effective training strategies to acquire essential task-specific skills from the low-resource data, rather than inversely reducing efficiency.

\begin{figure}[h]
\begin{center}
\includegraphics[width=10cm]{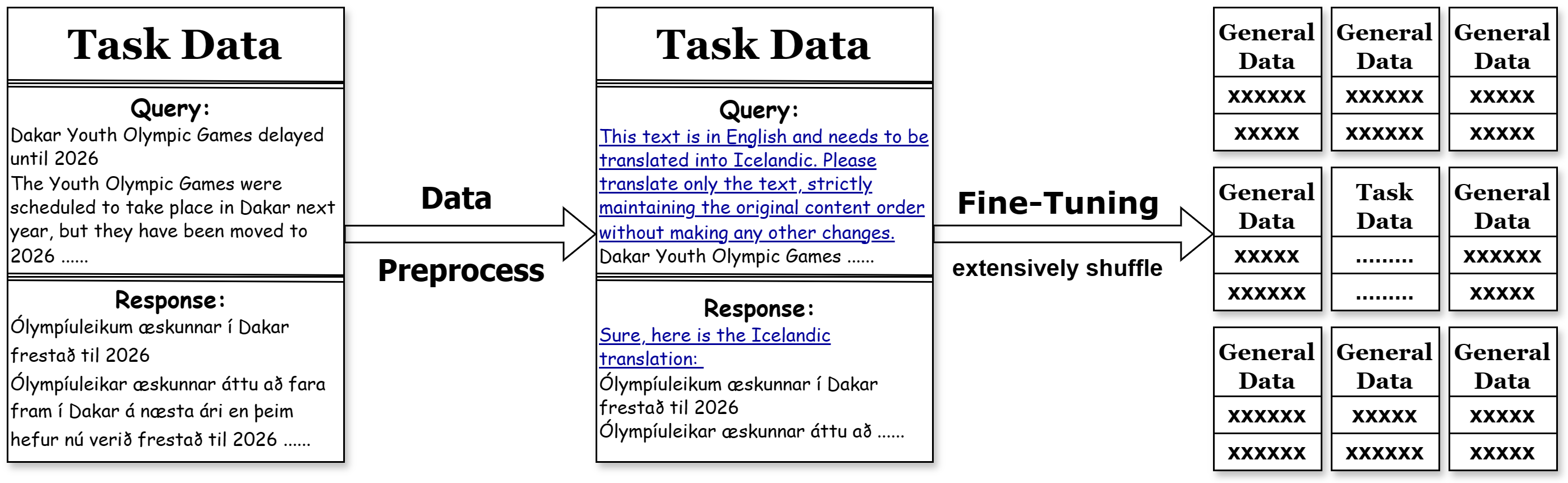}
\end{center}
\caption{An overview of Token-Efficient Leverage Learning}
\label{fig:TELLOverview}
\end{figure}

Our TELL strategy, depicted in Figure~\ref{fig:TELLOverview}, addresses the two issues by first employing data preprocessing, which enriches the low-resource dataset with an "anchor prompt" to implant consistent semantic features. Second, it shuffles a large amount of general data with the low-resource data to ensure a seamless integration of task-specific and general knowledge. We will show extensively shuffling is a effective training strategies to fit the issue later.

Our strategy deliberately avoids intricate design to underscore that the efficiency stems from the intrinsic nature of Leverage Learning, rather than from any sophisticated fine design on our part. The TELL strategy leverages only the data available in low-resource settings and open-source general datasets, circumventing the need for further augment or expansion of the low-resource datasets, thus preventing the introduction of unintended biases. By not incorporating additional context during inference, TELL avoids the unpredictable impact on model performance associated with in-context learning \citep{webson2022promptbased, zhao2021calibrate}. Our ablation studies confirm that these enhancements primarily arise from the synergistic effect of task and general data within the TELL method. For low-resource tasks at the scale of \(10^4\) tokens\footnote{In order to facilitate the same counting, llama2 tokenizer is used as a token counting tool in this article.}, direct application of SFT failed to yield noticeable performance improvements over the original model. However, the TELL strategy not only trains on these tasks but also achieves performance comparable to SFT fine-tuned on a task dataset nearly an order of magnitude larger. For tasks ranging between \(10^5\) to \(10^6\) tokens, TELL significantly outperforms SFT when trained on the same task data, again matching the performance of SFT on an expanded dataset nearly an order of magnitude larger. Across our experiments, TELL consistently showed superior performance-per-token improvement over traditional SFT. This strategy has been successful across various tasks and LLMs.

Our contributions are as follows:

\begin{enumerate}
    \item We introduce a novel fine-tuning methodology for low-resource tasks, termed Leverage Learning, and have developed a minimalist implementation, Token-Efficient Leverage Learning (TELL) to validate the efficacy of Leverage Learning. TELL significantly reduces the reliance on task-specific data while achieving performance comparable to fine-tuning on substantially larger task datasets. It also demonstrates a marked improvement in performance per task token over traditional SFT.
    \item Our strategy avoids the unpredictable impact on model performance associated with in-context learning, offering a new solution for LLM learning in low-resource scenarios.
    \item We elucidate the scaling phenomenon of LLM's task-specific capabilities with increasing amounts of general data in the TELL method, along with the underlying "emergent ability" phenomenon, and attempt to interpret these findings through the lens of quantization hypothesis.
\end{enumerate}

\section{Related Works}

\subsection{LLM Tuning Methods Across Different Data Volumes}

The approach to enhancing task performance with LLMs varies with the available training data and budget. Discussing a typical model with 7B parameters, when the available task training data exceeds \(10^9\) tokens, current studies predominantly employ continued pre-training to improve task performance \citep{Wen2023-qj}, resulting in task-specific LLMs for that domain. As the volume of task training data decreases to between \(10^7\) and \(10^9\) tokens, the focus shifts to fine-tuning LLMs, with instructional fine-tuning being the prevalent method. There are numerous open-source datasets with corresponding practice in this range, such as those benefiting mathematical reasoning \citep{Sun2023-rz} and code generation \citep{Lee2023-yd, Yu2023-vm}. Besides domain-specific tasks, open-source general datasets like \citet{alpaca} are extensively employed for training models on instruction alignment. When task training data further decreases to the \(10^6\) tokens range, simple SFT becomes challenging. \citet{Zhou2023-pz} has shown that high-quality, diverse data at this scale can yield favorable fine-tuning outcomes, at the meantime illustrating a clear quality disparity between LLM outputs trained on high and low-quality data. With a reduction to \(10^5\) tokens, direct SFT, regardless of employing PEFT, struggles to achieve satisfactory results, leading to potential underlearning of task features or overfitting, thereby degrading the model's general capabilities \citep{Aghajanyan2020-fr}, a situation that worsens in small-sample contexts \citep{Zhang2020-gp}. In this scenario, unless using the task data as seed prompts for reinforcement learning \citep{Lee2023-lo} through methods like DPO \citep{Rafailov2023-ir} and PPO \citep{Schulman2017-mt}, it becomes difficult to obtain good results, further escalating training costs. When data volume drops to \(10^4\) tokens, to the best of our knowledge, no studies have shown how to fine-tune LLMs effectively at this scale; in-context learning remains the only viable option \citep{Min2022-ei}, enabling few-shot learning through a series of examples or meta descriptions in the context. However, in-context learning has a series of shortcomings as discussed in the Introduction.

As a result, there exists a mismatch: the data volume for low-resource tasks in real-world scenarios often falls between \(10^4\) and \(10^6\) tokens \citep{wmt2021}, whereas existing LLM fine-tuning research predominantly targets the \(10^7\) to \(10^9\) token range. This discrepancy renders current research inapplicable for fine-tuning on low-resource tasks directly. Existing data augmentation methods are unsuitable for these scenarios due to the costs of human evaluation or privacy risks. Hence, there is a need for a new, token-efficient method that relies on existing task data and can effectively fine-tune LLMs for low-resource tasks, enabling them to acquire the necessary task capabilities.

\subsection{Mixed Fine-Tuning}

Extensively shuffling, a important aspect of our TELL strategy, leverages the abundance of general data to mitigate noise fluctuations, simplifying the problem. We briefly overview mixed task data fine-tuning, a prevalent technique in which the ratio of general to task-specific data is a key differentiator among studies.

Mixed task data fine-tuning involves varying ratios of general to task-specific data. Traditional fine-tuning uses 0\% general data, relying solely on task-specific datasets \citep{Chung2022-vk}. \citet{Chung2022-vk} indicates that incorporating diverse tasks in fine-tuning can enhance model performance. As tasks diversify, exemplified by \citet{Michaud2023-rs}, incorporating up to 1836 tasks, the prevalence of large-scale general data \citep{Dong2023-uv} in fine-tuning grows. 

While traditional practice ranges from 0\% to 100\% general data, concerns about semantic discrepancies have limited exploration of higher ratios. \citet{Dong2023-uv} explored ratios up to 200\%, finding increased general data led to performance fluctuations attributed to noise. Our TELL experiment, however, indicates that at "smaller" ratios (0\% to 1000\%), LLM performance aligns with \citet{Dong2023-uv}. Yet, employing Leverage Learning with higher ratios yields performance improvements for low-resource tasks.

\section{Leverage Learning and TELL}

The related work provides valuable insights into fine-tuning strategies across various data volumes and explores the nature of mixed fine-tuning. However, these studies have not extensively investigated large-scale mixed fine-tuning, nor have they successfully utilized mixed fine-tuning to address the challenges of fine-tuning for low-resource tasks. Our work addresses this gap. We introduce Leverage Learning, a methodology inspired by quantization hypothesis \citep{Michaud2023-rs}, and present a minimalist strategic implementation, Token-Efficient Leverage Learning (TELL), to demonstrate the effectiveness of Leverage Learning. Even with this minimalistic implementation of Leverage Learning, without intricate design, it significantly reduces reliance on task-specific data and achieves performance comparable to fine-tuning on large-scale task datasets.

\subsection{Leverage Learning}

Leverage Learning is a methodology with the core vision of fully utilizing information from low-resource task data. It aims for LLMs to acquire task-specific capabilities from such data while learning non-specific capabilities from general data. 

According to the hypothesis proposed in Michaud et al. (2023), LLMs operate on "quanta," fundamental units of capability, enabling them to generalize to unfamiliar tasks by leveraging commonalities between extensive training data and specific task requirements. This suggests that essential quanta can be extracted from general data for low-resource tasks, reducing reliance on task-specific datasets. Analogous to the lever principle in physics, Leverage Learning uses extensive general data to facilitate learning with minimal low-resource data, embodying the core concept of Leverage Learning.

Despite the promising vision, two pivotal issues emerge:

Firstly, humans naturally distinguish between low-resource and general data, leveraging prior knowledge, a skill LLMs often lack, which hinders their task-specific performance enhancement. Secondly, even if LLMs manage to differentiate these data types, a strategic training approach is needed for them to effectively learn task-specific skills from low-resource data and broader skills from general data. This contrasts with standard fine-tuning on large datasets, where LLMs acquire both sets of skills indiscriminately, which is precisely the scenario we aim to transcend.

We believe that addressing these two issues will enable us to achieve the vision of Leverage Learning. We show two universal solutions as Leverage Learning. For the former issue, implant semantic features within low-resource tasks could facilitate LLMs in differentiating them from general data during training. Regarding the latter, we deem that altering the training sequence could serve as a solution. We shall elucidate how this altered training sequence works and demonstrate that "extensively shuffle" is a special instance of this solution.

To clarify how adjusting the training sequence can address the challenge of efficiently learning from low-resource data, we focus on the core principles of the quantization hypothesis \citep{Michaud2023-rs}. This hypothesis suggests that abilities can be deconstructed into discrete units known as "quantas," which are either acquired or not, and the model's performance is intricately linked to these acquired quantas. Furthermore, quantas have a natural hierarchy in their utility for loss reduction, leading to an optimal learning sequence, termed the "Q Sequence."

Our analysis assume that some quanta are interdependent (denoted as \(q^c\)) and must be learned in sequence, while others (denoted as \(q^i\)) can be learned independently. For low-resource tasks, the Q sequence may initially include both \(q^i\) and \(q^c\), but with enough general and targeted task data, the sequence can evolve to primarily task feature \(q^c\) quanta related to task-specific features. If not, we can shift the unexpected quantas into the previous "enough and targeted" quanta set, then the situation loops back to our expectation.

The intuitive optimal training sequence for low-resource tasks is sequentially learning task-specific \(q^c\) quanta, followed by learning any remaining \(q^i\) quanta. The challenge lies in identifying these quanta and their order without direct observation, yet optimal utilization of low-resource data suggests a training pattern where task data is interspersed with general data, applied general data to protect task data from being wasted.

To operationalize this, we propose a pragmatic approach: randomly mix tasks and a lot of general data, ensuring a comprehensive shuffle. We call it "extensively shuffle". This method underpins the TELL strategy, optimizing the use of low-resource data and reflecting the essence of Leverage Learning.

\subsection{Components of TELL}

The TELL strategy, rooted in Leverage Learning, consists of two primary components: data preprocessing and model fine-tuning, aimed at optimizing the use of low-resource task data. In data preprocessing, "anchor prompts" are used to implant consistent semantic features into the low-resource data, helping LLMs distinguish it from general data during training. This approach enables LLMs to better leverage general data for learning specific tasks.

In model fine-tuning, the focus is on maximizing task-specific learning under tight budget constraints, often employing Parameter-Efficient Fine-Tuning (PEFT) methods like LoRA. A key tactic is to significantly increase the volume of general data and then integrate it with task data through extensive shuffling, enhancing the model's ability to learn from limited task-specific information.

Empirical evidence suggests that a very high ratio of general to task-specific data (typically beyond 1000\%) is crucial for improving performance on low-resource tasks, surpassing the conventional ratios used in multi-task learning scenarios. This approach has hinted at potential emergent improvements in LLM performance, significantly boosting their performance on these tasks. While the current method of random mixing serves to validate the effectiveness of Leverage Learning, we believe that further refined strategies could enhance its performance even more.

\section{Experiments}

\subsection{Experimental Setup}

We developed two datasets for fine-tuning LLMs: Halu-CoT-JSON, designed for fine-tuning LLMs' capabilities in following structured text formats, and IS-EN-NEWS, designed for fine-tuning LLMs in low-resource language translation.\footnote{Due to length distribution differences between the WMT-21 and alpaca-en datasets, we use entity ratios in our comparative experiments to better reflect model performance in relation to training data.} The detailed methodology for constructing these datasets is elaborated in the appendix.

Experiments were conducted on three open-source models: LLaMA2-7B-Chat \citep{Touvron2023-wn}, QWen-7B-Chat \citep{Bai2023-jy}, and Gemma-7b-it \citep{gemma}. We fine-tuned the first two models using TELL with the Halu-CoT-JSON dataset to test LLMs' capabilities in adhering to structured text formats. The Gemma-7b-it model was fine-tuned using TELL with the IS-EN-NEWS dataset to evaluate LLMs' abilities in low-resource language translation. The general dataset used for fine-tuning was alpaca-en\citep{alpaca}, and testing was conducted on benchmarks relevant to the respective domains. We set up experiments like this to verify the effect of TELL on the same task but with different models and the same model and different tasks.

We conducted comparative experiments with Instruction Supervised Fine-tuning (SFT), introduced by \citet{Chung2022-vk}, to validate the effectiveness of TELL. The "SFT" mentioned in the experiment refers to training using LoRA\citep{hu2021lora} with typical methods. LoRA and TELL are orthogonal. LoRA has been reported to possess good capabilities for small-sample learning \citep{Sun2023-rz,hu2021lora}, hence serving as a strong baseline. To control variables, we also applied the TELL strategy on LoRA. Training details are provided in the appendix.

\subsection{Experimental Results}

\subsubsection{Can TELL Solve the First Issue of Leverage Learning?}

The first issue of Leverage Learning refers to the challenge LLMs face in differentiating between low-resource task data and general data during training. The TELL strategy attempts to address this issue by introducing the concept of "anchor prompt". To verify the effectiveness of incorporating an anchor prompt, we conducted experiments using the llama2-7b-chat model on the IS-EN-NEWS dataset and the alpaca-en dataset. The goal of our experiments was to assess how the inclusion of an anchor prompt affects the semantic distance between task data queries and general data queries.

We extracted 50 entities from the IS-EN-NEWS dataset for Icelandic-to-English and English-to-Icelandic translations, processed with and without an anchor prompt in the llama2-7b-chat model, and visualized the 16th layer's representations using t-SNE, which is the intermediate layer of this model.

\begin{figure}[h]
  \centering
  \begin{minipage}{0.45\textwidth}
    \centering
    \includegraphics[width=5.7cm]{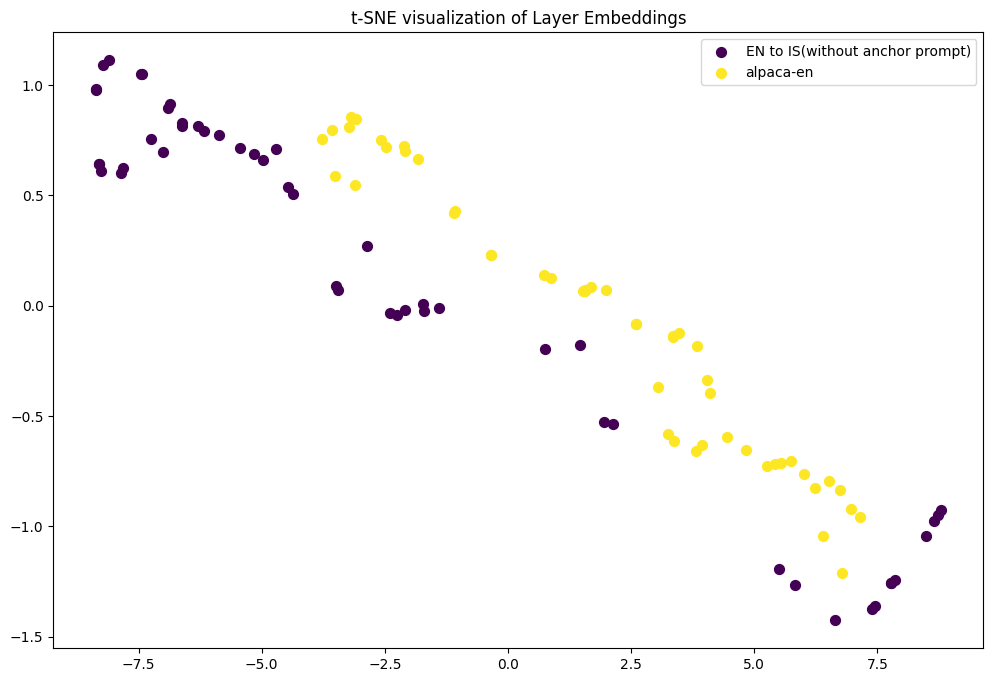}
    \caption{t-SNE visualization without anchor prompt}
    \label{fig:prefix_comparison1}
  \end{minipage}
  \hfill
  \begin{minipage}{0.45\textwidth}
    \centering
    \includegraphics[width=5.7cm]{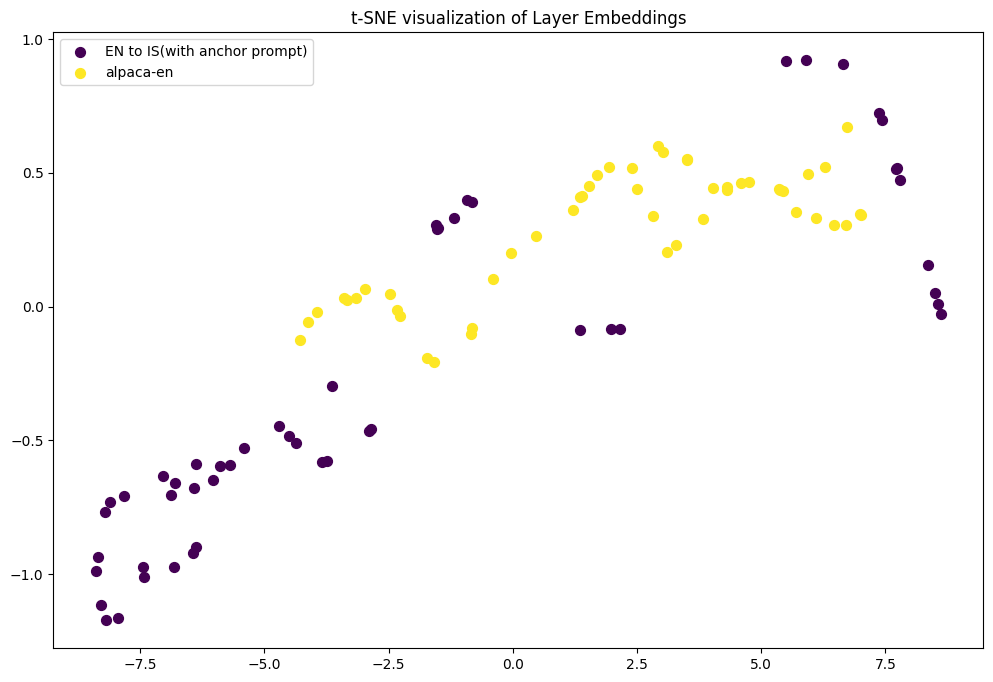}
    \caption{t-SNE visualization with anchor prompt}
    \label{fig:prefix_comparison2}
  \end{minipage}
\end{figure}

Visualization showed that without the anchor prompt, the 16th layer's semantic distributions of task and general data queries were alike. After applying the anchor prompt, however, the semantic distance increased, indicating distinct distribution patterns. This implies that the anchor prompt helps LLMs better distinguish between task and general data. As shown in Figure~\ref{fig:prefix_comparison1} and Figure~\ref{fig:prefix_comparison2}.

\begin{figure}[h]
  \centering
  \begin{minipage}{0.45\textwidth}
    \centering
    \includegraphics[width=5.7cm]{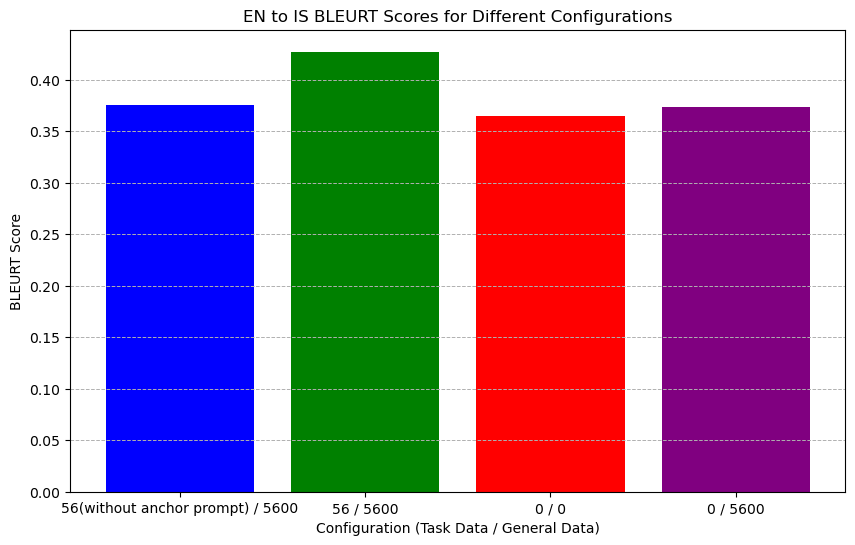}
    \caption{EN to IS BLEURT Scores with anchor prompt or not}
    \label{fig:prefix_comparison_bleurt1}
  \end{minipage}
  \hfill
  \begin{minipage}{0.45\textwidth}
    \centering
    \includegraphics[width=5.7cm]{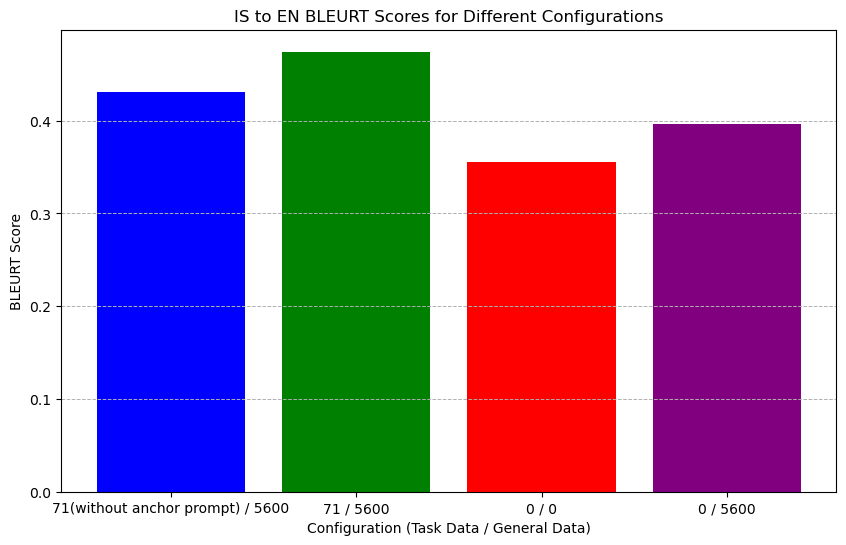}
    \caption{IS to EN BLEURT Scores with anchor prompt or not}
    \label{fig:prefix_comparison_bleurt2}
  \end{minipage}
\end{figure}

To confirm if this semantic difference impacts performance, we tested on the IS-EN-NEWS dataset with and without anchor prompts. Results showed that the anchor prompt significantly enhanced BLEURT scores in TELL, affirming that the TELL strategy effectively meets Leverage Learning’s primary requirement by improving LLMs' ability to differentiate task data from general data. As shown in Figure~\ref{fig:prefix_comparison_bleurt1} and Figure~\ref{fig:prefix_comparison_bleurt2}.

\subsubsection{Ablation Study}
\label{ablation}

To ascertain that the improvements brought by the TELL strategy stemmed from the synergistic effect of task and general data, rather than merely the enhancement of general data, we conducted an ablation study. This study involved testing the Halu-CoT-JSON dataset on the llama2-7b-chat and QWen-7b-chat models, and the IS-EN-NEWS dataset on the Gemma-7b-it model. The task tokens for the former were \(10^6\) and for the latter, \(10^5\). Our findings indicate that TELL's performance was significantly better compared to using only general or task data. Additionally, TELL converged faster than SFT, requiring less data to achieve similar performance.

\begin{figure}[h]
  \centering
  \begin{minipage}{0.45\textwidth}
    \centering
    \includegraphics[width=5.7cm]{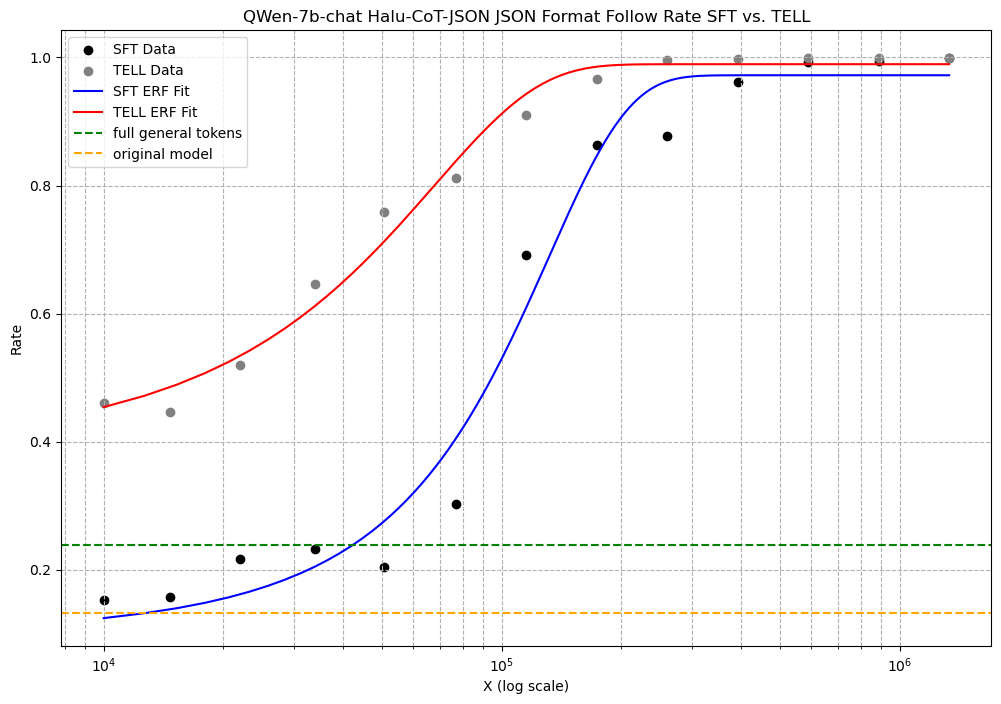}
    \caption{QWen-7b-chat Halu-CoT-JSON JSON format adherence rate}
    \label{fig:qwen_json}
  \end{minipage}
  \hfill
  \begin{minipage}{0.45\textwidth}
    \centering
    \includegraphics[width=5.7cm]{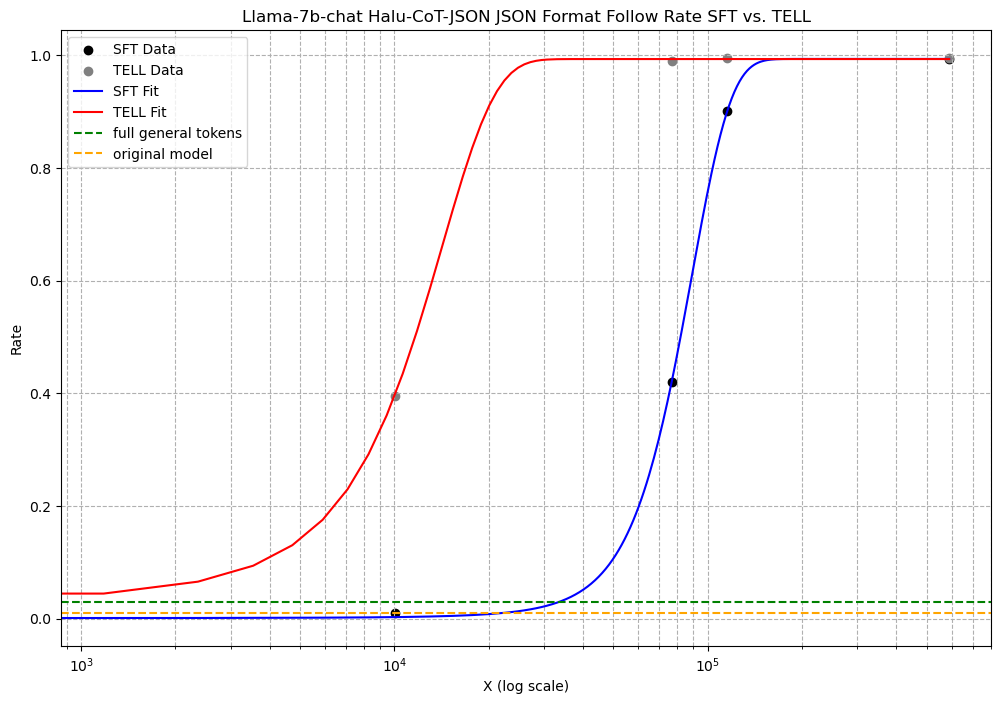}
    \caption{llama2-7b-chat Halu-CoT-JSON JSON format adherence rate}
    \label{fig:llama2_json}
  \end{minipage}
\end{figure}

Experimental results on the QWen-7b-chat model demonstrated that within the range of \(10^4\) to \(10^6\) task tokens, TELL consistently outperformed SFT. A similar phenomenon was observed in the llama-7b-chat model. Both methods eventually converged to a 100\% JSON format adherence rate. The experiment shows that TELL enabled LLMs to learn capabilities for low-resource tasks from the additional general data. As shown in Figure~\ref{fig:qwen_json} and Figure~\ref{fig:llama2_json}.

As a benchmark, the performance of fine-tuning with only general data or the original LLM is depicted as dashed lines in the graphs. We observed that in the TELL approach, the LLM's enhanced capability from general data—namely, the performance boost TELL provided over SFT for an equivalent amount of task tokens—was significantly greater than the enhancement from using only general data on the original LLM. This indicates that TELL's performance improvement primarily results from the synergistic effect of general and task data.

\begin{figure}[h]
  \centering
  \begin{minipage}{0.45\textwidth}
    \centering
    \includegraphics[width=5.7cm]{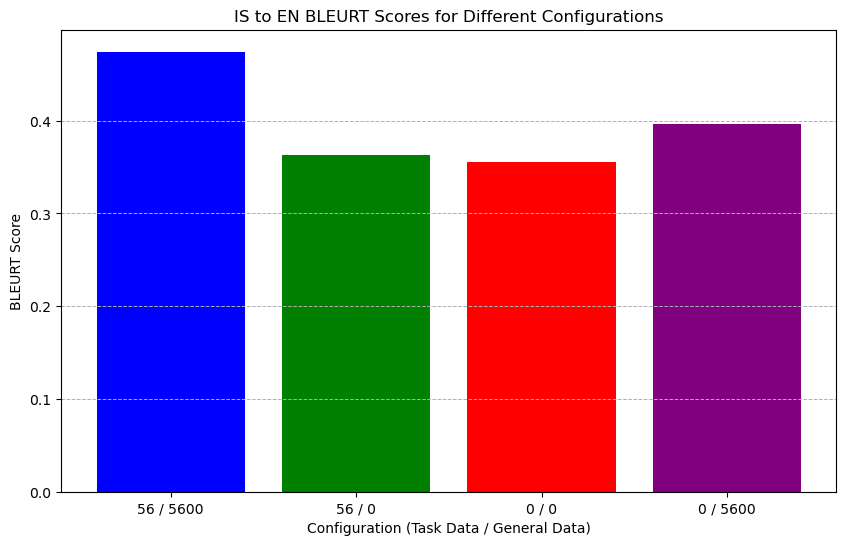}
    \caption{Gemma-7b-it Icelandic-to-English translation BLEURT score}
    \label{fig:is-en1}
  \end{minipage}
  \hfill
  \begin{minipage}{0.45\textwidth}
    \centering
    \includegraphics[width=5.7cm]{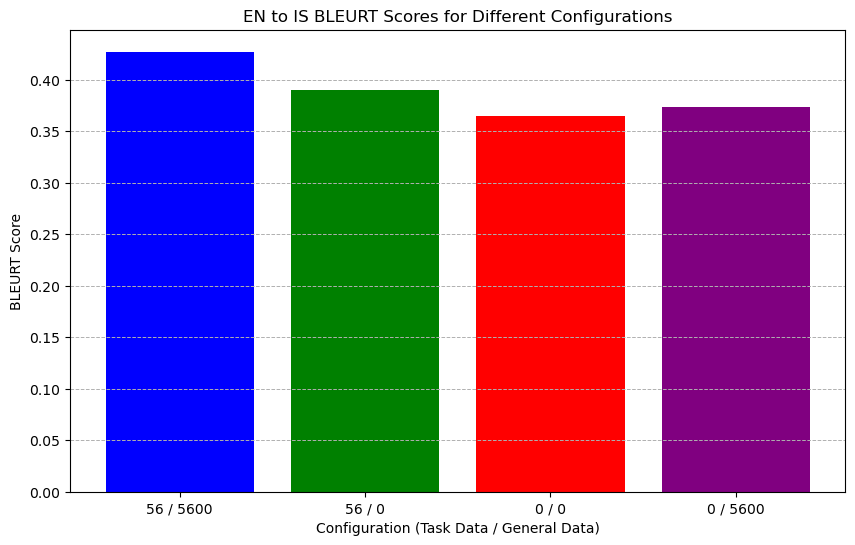}
    \caption{Gemma-7b-it English-to-Icelandic translation BLEURT score}
    \label{fig:en-is1}
  \end{minipage}
\end{figure}

In the Icelandic-to-English translation task, due to the small size of the dataset, we did not test TELL's performance scaling with data volume. Instead, we conducted an ablation study using only the WMT-21 subset of the dataset. We observed that TELL's performance improvement on this low-resource task was notably stronger than that achieved by using only general or task data. This suggests that for the Icelandic-to-English translation task, TELL's performance improvement principally arises from the synergistic effect of general and task data. As shown in Figure~\ref{fig:is-en1} and Figure~\ref{fig:en-is1}.

Similar phenomena were validated in the ablation study for the English-to-Icelandic translation task.

\subsubsection{Leverage Learning Scaling with General Tokens}

The aforementioned experiments validated the performance enhancement and high token-efficiency of the TELL strategy in low-resource tasks, investigating the converged performance of the TELL approach. To further comprehend the behavior of TELL's performance variation, we delved into how its performance scales with general data when it is not yet converged, particularly with lesser amounts of general data.

Consistent with the previous experimental setup, we designed two experiments using two subsets of the Halu-CoT-JSON dataset, containing 15 entries (1e4 tokens) and 115 entries (7e4 tokens), respectively. These data points were chosen because the performance achieved with 1e4 tokens using the TELL strategy approximates that of directly performing SFT with 7e4 tokens as shown in Experiment \ref{ablation}. These experiments aimed to deepen our understanding of TELL and confirm that the TELL strategy maximizes the model performance improvement potential from an equivalent volume of data compared to SFT.

\begin{figure}[h]
  \centering
  \begin{minipage}{0.45\textwidth}
    \centering
    \includegraphics[width=5.7cm]{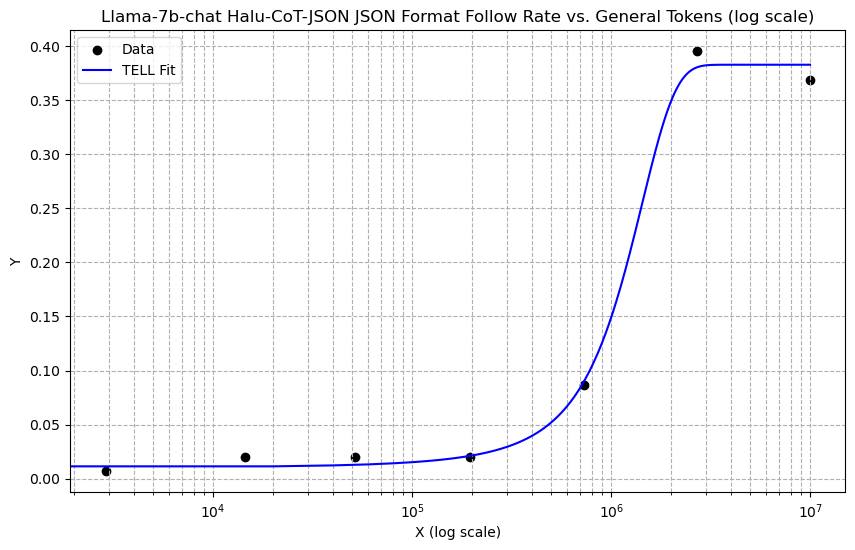}
    \caption{1e4 task tokens, Leverage Learning scaling with general tokens}
    \label{fig:1e4ll}
  \end{minipage}
  \hfill
  \begin{minipage}{0.45\textwidth}
    \centering
    \includegraphics[width=5.7cm]{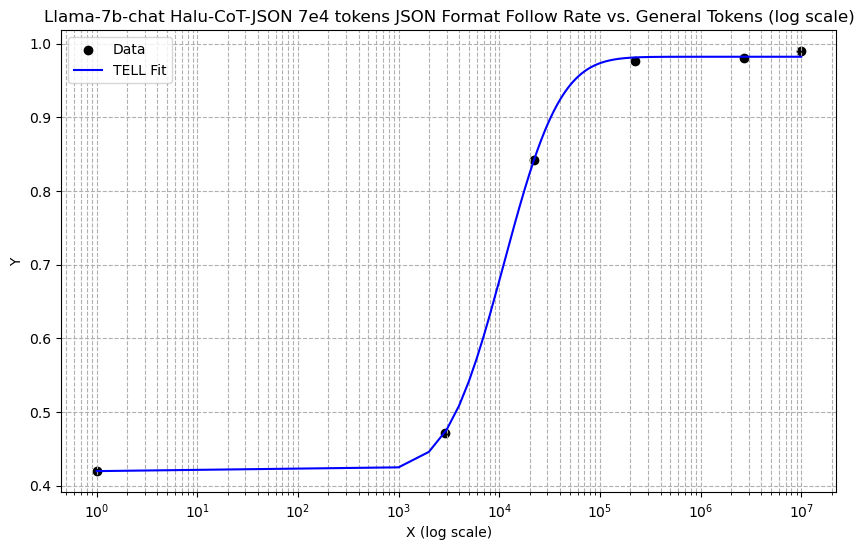}
    \caption{7e4 task tokens, Leverage Learning scaling with general tokens}
    \label{fig:7e4ll}
  \end{minipage}
\end{figure}

We observed a phenomenon similar to "emergent ability" in the performance change of LLMs on low-resource tasks with increasing general data ratios. At lower general data ratios, no significant improvement in task performance was noted, aligning with previous research findings. However, once the ratio expanded beyond a certain threshold, a rapid enhancement in performance was evident. In our experiments, for a fixed amount of task data, the LLM's performance eventually converged to a specific value. As shown in Figure~\ref{fig:1e4ll} and Figure~\ref{fig:7e4ll}.

It was noted that fine-tuning with just 15 entries (1e4 tokens) increased the JSON adherence accuracy from 1\% in the original model to 39.54\%. This result is close to the 42\% accuracy achieved by direct fine-tuning with 115 entries (7e4 tokens). When the fine-tuning data was increased to 115 entries, the TELL method further raised the JSON adherence accuracy to 99\%, comparable to the outcome of direct fine-tuning with 883 entries (6e5 tokens).

Experiments were also conducted on the WMT-21 subset of the IS-EN-NEWS dataset, testing two translation tasks between English and Icelandic. Similarly, it was observed that with the increase of general data, the performance of the gemma-7b-it model on the IS-EN-NEWS dataset using the SFT method with all subset data was far inferior to the TELL strategy. Even after supplementing with the subset described in the appendix from the University of Iceland's TM, the BLEURT scores obtained with SFT, with 5 times more data available for the low-resource task than TELL, still fell short.

\begin{figure}[h]
  \centering
  \begin{minipage}{0.45\textwidth}
    \centering
    \includegraphics[width=5.7cm]{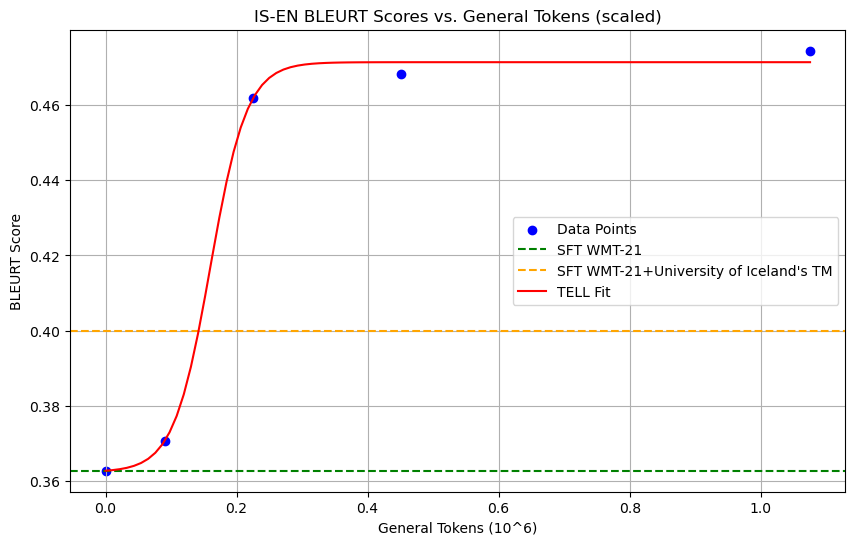}
    \caption{IS to EN, Leverage Learning scaling with general tokens}
    \label{fig:is-en2}
  \end{minipage}
  \hfill
  \begin{minipage}{0.45\textwidth}
    \centering
    \includegraphics[width=5.7cm]{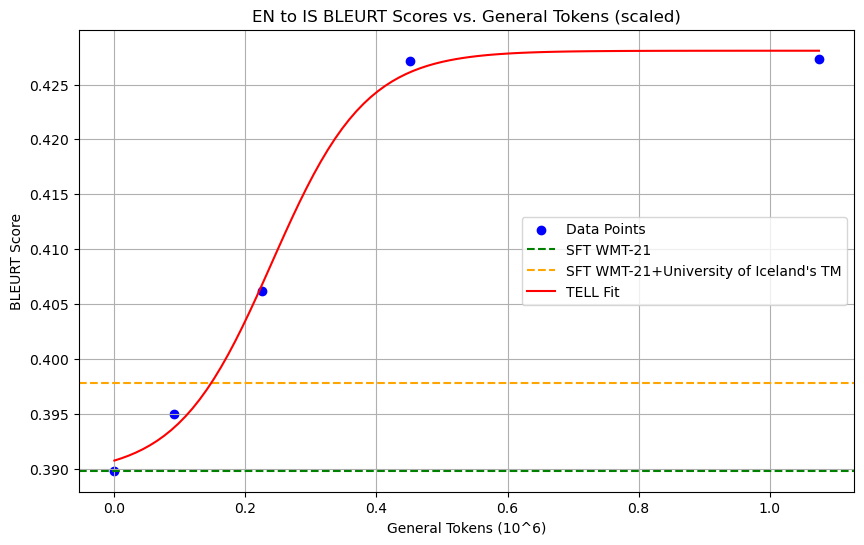}
    \caption{EN to IS, Leverage Learning scaling with general tokens}
    \label{fig:en-is2}
  \end{minipage}
\end{figure}

The JSON Format Follow is considered a discontinuous metric, while BLEURT is widely regarded as a continuous one. Thus, the accuracy of LLMs during learning follows an S-curve as general data increases, irrespective of the metrics' linearity or continuity. However, according to \citet{Schaeffer2023-ts}, phenomena like the "emergent ability" of nonlinear capability enhancements typically manifest in nonlinear or discontinuous metrics. Therefore, the occurrence of emergence in BLEURT metrics suggests potential limitations in the research's interpretation. We discuss the interpretation of the emergence phenomenon in the appendix section, grounded on Leverage Learning. As shown in Figure~\ref{fig:is-en2} and Figure~\ref{fig:en-is2}.

\section{Conclusion}

We introduced Leverage Learning, a novel methodology to enhance LLM performance on low-resource tasks, validated through the TELL strategy, which efficiently combines anchor prompts with scaled-up general data. Our experiments across various LLMs and datasets demonstrated TELL's superior performance and token-efficiency over traditional SFT, highlighting the synergistic effect of general and task-specific data. The 'emergent ability' phenomenon observed in TELL, along with potential cost reductions through self-generated anchor prompts, suggests promising avenues for fine-tuning LLMs in low-resource scenarios.

\bibliography{Token_Efficient_Leverage_Learning_in_Large_Language_Model}

\begin{thebibliography}{35}
\providecommand{\natexlab}[1]{#1}
\providecommand{\url}[1]{\texttt{#1}}
\expandafter\ifx\csname urlstyle\endcsname\relax
  \providecommand{\doi}[1]{doi: #1}\else
  \providecommand{\doi}{doi: \begingroup \urlstyle{rm}\Url}\fi

\bibitem[Aghajanyan et~al.(2021)Aghajanyan, Shrivastava, Gupta, Goyal,
  Zettlemoyer, and Gupta]{Aghajanyan2020-fr}
Armen Aghajanyan, Akshat Shrivastava, Anchit Gupta, Naman Goyal, Luke
  Zettlemoyer, and Sonal Gupta.
\newblock Better fine-tuning by reducing representational collapse.
\newblock In \emph{Proc. of ICLR}. OpenReview.net, 2021.
\newblock URL \url{https://openreview.net/forum?id=OQ08SN70M1V}.

\bibitem[Ass(2021)]{wmt2021}
\emph{Proceedings of the 6th Conference on Machine Translation}, Location of
  the Conference, 2021. Association for Computational Linguistics.

\bibitem[Bai et~al.(2023)Bai, Bai, Chu, Cui, Dang, Deng, Fan, Ge, Han, Huang,
  Hui, Ji, Li, Lin, Lin, Liu, Liu, Lu, Lu, Ma, Men, Ren, Ren, Tan, Tan, Tu,
  Wang, Wang, Wang, Wu, Xu, Xu, Yang, Yang, Yang, Yang, Yao, Yu, Yuan, Yuan,
  Zhang, Zhang, Zhang, Zhang, Zhou, Zhou, Zhou, and Zhu]{Bai2023-jy}
Jinze Bai, Shuai Bai, Yunfei Chu, Zeyu Cui, Kai Dang, Xiaodong Deng, Yang Fan,
  Wenbin Ge, Yu~Han, Fei Huang, Binyuan Hui, Luo Ji, Mei Li, Junyang Lin, Runji
  Lin, Dayiheng Liu, Gao Liu, Chengqiang Lu, Keming Lu, Jianxin Ma, Rui Men,
  Xingzhang Ren, Xuancheng Ren, Chuanqi Tan, Sinan Tan, Jianhong Tu, Peng Wang,
  Shijie Wang, Wei Wang, Shengguang Wu, Benfeng Xu, Jin Xu, An~Yang, Hao Yang,
  Jian Yang, Shusheng Yang, Yang Yao, Bowen Yu, Hongyi Yuan, Zheng Yuan,
  Jianwei Zhang, Xingxuan Zhang, Yichang Zhang, Zhenru Zhang, Chang Zhou,
  Jingren Zhou, Xiaohuan Zhou, and Tianhang Zhu.
\newblock Qwen technical report.
\newblock 2023.

\bibitem[Chung et~al.(2022)Chung, Hou, Longpre, Zoph, Tay, Fedus, Li, Wang,
  Dehghani, Brahma, Webson, Gu, Dai, Suzgun, Chen, Chowdhery, Castro-Ros,
  Pellat, Robinson, Valter, Narang, Mishra, Yu, Zhao, Huang, Dai, Yu, Petrov,
  Chi, Dean, Devlin, Roberts, Zhou, Le, and Wei]{Chung2022-vk}
Hyung~Won Chung, Le~Hou, Shayne Longpre, Barret Zoph, Yi~Tay, William Fedus,
  Yunxuan Li, Xuezhi Wang, Mostafa Dehghani, Siddhartha Brahma, Albert Webson,
  Shixiang~Shane Gu, Zhuyun Dai, Mirac Suzgun, Xinyun Chen, Aakanksha
  Chowdhery, Alex Castro-Ros, Marie Pellat, Kevin Robinson, Dasha Valter,
  Sharan Narang, Gaurav Mishra, Adams Yu, Vincent Zhao, Yanping Huang, Andrew
  Dai, Hongkun Yu, Slav Petrov, Ed~H Chi, Jeff Dean, Jacob Devlin, Adam
  Roberts, Denny Zhou, Quoc~V Le, and Jason Wei.
\newblock Scaling instruction-finetuned language models.
\newblock 2022.

\bibitem[Dettmers et~al.(2023)Dettmers, Pagnoni, Holtzman, and
  Zettlemoyer]{Dettmers2023-gh}
Tim Dettmers, Artidoro Pagnoni, Ari Holtzman, and Luke Zettlemoyer.
\newblock {QLoRA}: Efficient finetuning of quantized {LLMs}.
\newblock 2023.

\bibitem[Dong et~al.(2023)Dong, Yuan, Lu, Li, Xue, Liu, Wang, Yuan, Zhou, and
  Zhou]{Dong2023-uv}
Guanting Dong, Hongyi Yuan, Keming Lu, Chengpeng Li, Mingfeng Xue, Dayiheng
  Liu, Wei Wang, Zheng Yuan, Chang Zhou, and Jingren Zhou.
\newblock How abilities in large language models are affected by supervised
  fine-tuning data composition.
\newblock 2023.

\bibitem[Garcia et~al.(2023{\natexlab{a}})Garcia, Bansal, Cherry, Foster,
  Krikun, Feng, Johnson, and Firat]{Garcia2023-dx}
Xavier Garcia, Yamini Bansal, Colin Cherry, George Foster, Maxim Krikun,
  Fangxiaoyu Feng, Melvin Johnson, and Orhan Firat.
\newblock The unreasonable effectiveness of few-shot learning for machine
  translation.
\newblock 2023{\natexlab{a}}.

\bibitem[Garcia et~al.(2023{\natexlab{b}})Garcia, Bansal, Cherry, Foster,
  Krikun, Feng, Johnson, and Firat]{Garcia2023-qq}
Xavier Garcia, Yamini Bansal, Colin Cherry, George Foster, Maxim Krikun,
  Fangxiaoyu Feng, Melvin Johnson, and Orhan Firat.
\newblock The unreasonable effectiveness of few-shot learning for machine
  translation.
\newblock 2023{\natexlab{b}}.

\bibitem[Google(2024)]{gemma}
Google.
\newblock Gemma-7b-it.
\newblock \url{https://huggingface.co/google/gemma-7b-it}, 2024.

\bibitem[Ho et~al.(2022)Ho, Schmid, and Yun]{Ho2022-hz}
Namgyu Ho, Laura Schmid, and Se-Young Yun.
\newblock Large language models are reasoning teachers.
\newblock 2022.

\bibitem[Hu et~al.(2021)Hu, Shen, Wallis, Allen-Zhu, Li, Wang, Wang, and
  Chen]{hu2021lora}
Edward~J. Hu, Yelong Shen, Phillip Wallis, Zeyuan Allen-Zhu, Yuanzhi Li, Shean
  Wang, Lu~Wang, and Weizhu Chen.
\newblock Lora: Low-rank adaptation of large language models, 2021.

\bibitem[Kim et~al.(2022)Kim, Oh, Kim, and Yun]{Kim2022-hy}
Yujin Kim, Jaehoon Oh, Sungnyun Kim, and Se-Young Yun.
\newblock How to fine-tune models with few samples: Update, data augmentation,
  and test-time augmentation.
\newblock 2022.

\bibitem[Lee et~al.(2023{\natexlab{a}})Lee, Hunter, and Ruiz]{Lee2023-yd}
Ariel~N Lee, Cole~J Hunter, and Nataniel Ruiz.
\newblock Platypus: Quick, cheap, and powerful refinement of {LLMs}.
\newblock 2023{\natexlab{a}}.

\bibitem[Lee et~al.(2023{\natexlab{b}})Lee, Phatale, Mansoor, Lu, Mesnard,
  Bishop, Carbune, and Rastogi]{Lee2023-lo}
Harrison Lee, Samrat Phatale, Hassan Mansoor, Kellie Lu, Thomas Mesnard, Colton
  Bishop, Victor Carbune, and Abhinav Rastogi.
\newblock {RLAIF}: Scaling reinforcement learning from human feedback with {AI}
  feedback.
\newblock 2023{\natexlab{b}}.

\bibitem[Li et~al.(2023)Li, Cheng, Zhao, Nie, and Wen]{Li2023-wl}
Junyi Li, Xiaoxue Cheng, Wayne~Xin Zhao, Jian-Yun Nie, and Ji-Rong Wen.
\newblock {HaluEval}: A large-scale hallucination evaluation benchmark for
  large language models.
\newblock 2023.

\bibitem[Michaud et~al.(2023)Michaud, Liu, Girit, and Tegmark]{Michaud2023-rs}
Eric~J Michaud, Ziming Liu, Uzay Girit, and Max Tegmark.
\newblock The quantization model of neural scaling.
\newblock 2023.

\bibitem[Mihaylov et~al.(2018)Mihaylov, Clark, Khot, and
  Sabharwal]{Mihaylov2018-il}
Todor Mihaylov, Peter Clark, Tushar Khot, and Ashish Sabharwal.
\newblock Can a suit of armor conduct electricity? a new dataset for open book
  question answering.
\newblock In \emph{Proc. of EMNLP}, pp.\  2381--2391, Brussels, Belgium, 2018.
  Association for Computational Linguistics.
\newblock \doi{10.18653/v1/D18-1260}.
\newblock URL \url{https://aclanthology.org/D18-1260}.

\bibitem[Min et~al.(2022)Min, Lyu, Holtzman, Artetxe, Lewis, Hajishirzi, and
  Zettlemoyer]{Min2022-ei}
Sewon Min, Xinxi Lyu, Ari Holtzman, Mikel Artetxe, Mike Lewis, Hannaneh
  Hajishirzi, and Luke Zettlemoyer.
\newblock Rethinking the role of demonstrations: What makes in-context learning
  work?
\newblock In \emph{Proc. of EMNLP}, pp.\  11048--11064, Abu Dhabi, United Arab
  Emirates, 2022. Association for Computational Linguistics.
\newblock URL \url{https://aclanthology.org/2022.emnlp-main.759}.

\bibitem[OpenAI(2023)]{OpenAI2023GPT35Turbo}
OpenAI.
\newblock Gpt-3.5 turbo.
\newblock \url{https://openai.com/}, 2023.

\bibitem[Pu et~al.(2021)Pu, Chung, Parikh, Gehrmann, and Sellam]{Pu2021-tj}
Amy Pu, Hyung~Won Chung, Ankur Parikh, Sebastian Gehrmann, and Thibault Sellam.
\newblock Learning compact metrics for {MT}.
\newblock In \emph{Proc. of EMNLP}, pp.\  751--762, Online and Punta Cana,
  Dominican Republic, 2021. Association for Computational Linguistics.
\newblock \doi{10.18653/v1/2021.emnlp-main.58}.
\newblock URL \url{https://aclanthology.org/2021.emnlp-main.58}.

\bibitem[Rafailov et~al.(2023)Rafailov, Sharma, Mitchell, Ermon, Manning, and
  Finn]{Rafailov2023-ir}
Rafael Rafailov, Archit Sharma, Eric Mitchell, Stefano Ermon, Christopher~D
  Manning, and Chelsea Finn.
\newblock Direct preference optimization: Your language model is secretly a
  reward model.
\newblock 2023.

\bibitem[Schaeffer et~al.(2023)Schaeffer, Miranda, and
  Koyejo]{Schaeffer2023-ts}
Rylan Schaeffer, Brando Miranda, and Sanmi Koyejo.
\newblock Are emergent abilities of large language models a mirage?
\newblock 2023.

\bibitem[Schulman et~al.(2017)Schulman, Wolski, Dhariwal, Radford, and
  Klimov]{Schulman2017-mt}
John Schulman, Filip Wolski, Prafulla Dhariwal, Alec Radford, and Oleg Klimov.
\newblock Proximal policy optimization algorithms.
\newblock 2017.

\bibitem[Sun et~al.(2023)Sun, Ji, Ma, and Li]{Sun2023-rz}
Xianghui Sun, Yunjie Ji, Baochang Ma, and Xiangang Li.
\newblock A comparative study between full-parameter and {LoRA-based}
  fine-tuning on chinese instruction data for instruction following large
  language model.
\newblock 2023.

\bibitem[Taori et~al.(2023)Taori, Gulrajani, Zhang, Dubois, Li, Guestrin,
  Liang, and Hashimoto]{alpaca}
Rohan Taori, Ishaan Gulrajani, Tianyi Zhang, Yann Dubois, Xuechen Li, Carlos
  Guestrin, Percy Liang, and Tatsunori~B. Hashimoto.
\newblock Stanford alpaca: An instruction-following llama model.
\newblock \url{https://github.com/tatsu-lab/stanford_alpaca}, 2023.

\bibitem[Team(2024)]{swift}
The~ModelScope Team.
\newblock Swift:scalable lightweight infrastructure for fine-tuning.
\newblock \url{https://github.com/modelscope/swift}, 2024.

\bibitem[Touvron et~al.(2023)Touvron, Martin, Stone, Albert, Almahairi, Babaei,
  Bashlykov, Batra, Bhargava, Bhosale, Bikel, Blecher, Ferrer, Chen, Cucurull,
  Esiobu, Fernandes, Fu, Fu, Fuller, Gao, Goswami, Goyal, Hartshorn, Hosseini,
  Hou, Inan, Kardas, Kerkez, Khabsa, Kloumann, Korenev, Koura, Lachaux, Lavril,
  Lee, Liskovich, Lu, Mao, Martinet, Mihaylov, Mishra, Molybog, Nie, Poulton,
  Reizenstein, Rungta, Saladi, Schelten, Silva, Smith, Subramanian, Tan, Tang,
  Taylor, Williams, Kuan, Xu, Yan, Zarov, Zhang, Fan, Kambadur, Narang,
  Rodriguez, Stojnic, Edunov, and Scialom]{Touvron2023-wn}
Hugo Touvron, Louis Martin, Kevin Stone, Peter Albert, Amjad Almahairi, Yasmine
  Babaei, Nikolay Bashlykov, Soumya Batra, Prajjwal Bhargava, Shruti Bhosale,
  Dan Bikel, Lukas Blecher, Cristian~Canton Ferrer, Moya Chen, Guillem
  Cucurull, David Esiobu, Jude Fernandes, Jeremy Fu, Wenyin Fu, Brian Fuller,
  Cynthia Gao, Vedanuj Goswami, Naman Goyal, Anthony Hartshorn, Saghar
  Hosseini, Rui Hou, Hakan Inan, Marcin Kardas, Viktor Kerkez, Madian Khabsa,
  Isabel Kloumann, Artem Korenev, Punit~Singh Koura, Marie-Anne Lachaux,
  Thibaut Lavril, Jenya Lee, Diana Liskovich, Yinghai Lu, Yuning Mao, Xavier
  Martinet, Todor Mihaylov, Pushkar Mishra, Igor Molybog, Yixin Nie, Andrew
  Poulton, Jeremy Reizenstein, Rashi Rungta, Kalyan Saladi, Alan Schelten, Ruan
  Silva, Eric~Michael Smith, Ranjan Subramanian, Xiaoqing~Ellen Tan, Binh Tang,
  Ross Taylor, Adina Williams, Jian~Xiang Kuan, Puxin Xu, Zheng Yan, Iliyan
  Zarov, Yuchen Zhang, Angela Fan, Melanie Kambadur, Sharan Narang, Aurelien
  Rodriguez, Robert Stojnic, Sergey Edunov, and Thomas Scialom.
\newblock Llama 2: Open foundation and fine-tuned chat models.
\newblock 2023.

\bibitem[Webson \& Pavlick(2022)Webson and Pavlick]{webson2022promptbased}
Albert Webson and Ellie Pavlick.
\newblock Do prompt-based models really understand the meaning of their
  prompts?
\newblock In \emph{Proceedings of the 2022 Conference of the North American
  Chapter of the Association for Computational Linguistics: Human Language
  Technologies}, pp.\  2300--2344, Seattle, United States, 2022. Association
  for Computational Linguistics.
\newblock \doi{10.18653/v1/2022.naacl-main.167}.
\newblock URL \url{https://aclanthology.org/2022.naacl-main.167}.

\bibitem[Wei et~al.(2022)Wei, Wang, Schuurmans, Bosma, Ichter, Xia, Chi, Le,
  and Zhou]{Wei2022-kq}
Jason Wei, Xuezhi Wang, Dale Schuurmans, Maarten Bosma, Brian Ichter, Fei Xia,
  Ed~Chi, Quoc Le, and Denny Zhou.
\newblock Chain-of-thought prompting elicits reasoning in large language
  models.
\newblock 2022.

\bibitem[Wen et~al.(2023)Wen, Sun, Zhao, Fang, Chen, and Zou]{Wen2023-qj}
Cheng Wen, Xianghui Sun, Shuaijiang Zhao, Xiaoquan Fang, Liangyu Chen, and Wei
  Zou.
\newblock {ChatHome}: Development and evaluation of a domain-specific language
  model for home renovation.
\newblock 2023.

\bibitem[Yu et~al.(2023)Yu, Zhang, Shang, Huang, Xu, Zhao, Hu, and
  Yin]{Yu2023-vm}
Zhaojian Yu, Xin Zhang, Ning Shang, Yangyu Huang, Can Xu, Yishujie Zhao,
  Wenxiang Hu, and Qiufeng Yin.
\newblock {WaveCoder}: Widespread and versatile enhanced instruction tuning
  with refined data generation.
\newblock 2023.

\bibitem[Zhang et~al.(2021)Zhang, Wu, Katiyar, Weinberger, and
  Artzi]{Zhang2020-gp}
Tianyi Zhang, Felix Wu, Arzoo Katiyar, Kilian~Q. Weinberger, and Yoav Artzi.
\newblock Revisiting few-sample {BERT} fine-tuning.
\newblock In \emph{Proc. of ICLR}. OpenReview.net, 2021.
\newblock URL \url{https://openreview.net/forum?id=cO1IH43yUF}.

\bibitem[Zhao et~al.(2021)Zhao, Wallace, Feng, Klein, and
  Singh]{zhao2021calibrate}
Zihao Zhao, Eric Wallace, Shi Feng, Dan Klein, and Sameer Singh.
\newblock Calibrate before use: Improving few-shot performance of language
  models.
\newblock In Marina Meila and Tong Zhang (eds.), \emph{Proc. of ICML}, volume
  139 of \emph{Proceedings of Machine Learning Research}, pp.\  12697--12706.
  {PMLR}, 2021.
\newblock URL \url{http://proceedings.mlr.press/v139/zhao21c.html}.

\bibitem[Zhong et~al.(2023)Zhong, Cui, Guo, Liang, Lu, Wang, Saied, Chen, and
  Duan]{Zhong2023-um}
Wanjun Zhong, Ruixiang Cui, Yiduo Guo, Yaobo Liang, Shuai Lu, Yanlin Wang, Amin
  Saied, Weizhu Chen, and Nan Duan.
\newblock {AGIEval}: A human-centric benchmark for evaluating foundation
  models.
\newblock 2023.

\bibitem[Zhou et~al.(2023)Zhou, Liu, Xu, Iyer, Sun, Mao, Ma, Efrat, Yu, Yu,
  Zhang, Ghosh, Lewis, Zettlemoyer, and Levy]{Zhou2023-pz}
Chunting Zhou, Pengfei Liu, Puxin Xu, Srini Iyer, Jiao Sun, Yuning Mao, Xuezhe
  Ma, Avia Efrat, Ping Yu, Lili Yu, Susan Zhang, Gargi Ghosh, Mike Lewis, Luke
  Zettlemoyer, and Omer Levy.
\newblock {LIMA}: Less is more for alignment.
\newblock 2023.

\end{thebibliography}
\bibliographystyle{Token_Efficient_Leverage_Learning_in_Large_Language_Model}

\appendix
\section{Appendix}

\subsection{Training and Inference Details}

Unless specifically stated, all experimental data were obtained with a seed value of 42. For training, we employed the LoRA method as the Parameter-Efficient Fine-Tuning (PEFT) approach, with the following hyperparameters:
\begin{itemize}
    \item \texttt{lora\_rank}: 8
    \item \texttt{lora\_alpha}: 32
    \item \texttt{lora\_dropout\_p}: 0.05
    \item \texttt{optim}: \texttt{adamw\_torch}
    \item \texttt{learning\_rate}: 0.0001
    \item \texttt{weight\_decay}: 0.01
    \item \texttt{lr\_scheduler\_type}: \texttt{linear}
    \item \texttt{warmup\_ratio}: 0.03
    \item \texttt{temperature}: 0.3
    \item \texttt{top\_k}: 20
    \item \texttt{top\_p}: 0.7
    \item \texttt{repetition\_penalty}: 1.05
\end{itemize}

We employed the Swift training framework\cite{swift}, adhering to the recommendations from \citet{Dettmers2023-gh}, with a learning rate of \(10^{-4}\), batch size of 16, and due to evidence that most improvements occur in training phases of one epoch or less, we set epochs to 1.

For English-Icelandic translation tests, we used BLEURT scores, with the BLEURT-20 checkpoint and a fixed 2-shot learning approach, following the methodology described in \citet{Garcia2023-dx}.

When testing the Halu-CoT-JSON dataset, we used the unused portion of the HaluEval dataset as the test set. We adopted a 0-shot learning approach, consistent with the configuration used in practical hallucination detection systems.

Training was conducted using NVIDIA RTX 4090 GPUs. According to TensorBoard analytics, the consumed FLOPs ranged from 3e14 to 4e17, primarily due to the varying scale of the training data. In the training sets, the magnitude of task data ranged from \(10^4\) to \(10^6\) tokens, while for the general dataset, it varied from \(10^4\) to \(10^7\) tokens, depending on the experimental setup.

\subsection{Dataset Construction}

We developed two datasets, Halu-CoT-JSON and IS-EN-NEWS. The Halu-CoT-JSON dataset is designed for fine-tuning LLMs in text formatting adherence, while the IS-EN-NEWS dataset targets fine-tuning LLMs for low-resource language translation capabilities.

For the Halu-CoT-JSON dataset, we initially employed the gpt3.5-turbo-0613 model \citep{OpenAI2023GPT35Turbo} to generate thought chains \citep{Wei2022-kq} based on the HaluEval \citep{Li2023-wl} dataset, integrating these into a JSON format. This structured approach is intended to compel the model to engage in "thought" before determining if a hallucination is present, outputting the results in a format that is easily parseable. As a result, we created the Halu-CoT-JSON dataset.

Regarding the IS-EN-NEWS dataset, we utilized the training and test sets from the English-to-Icelandic and Icelandic-to-English translation tasks in the WMT-21 dataset. We also collect the training set with Icelandic-English bilingual translations longer than 50 words extracted from the University of Iceland's TM dataset, in order to boost performance for SFT. When we talk about the 'IS-EN-NEWS' dataset, if not mentioned explicitly about the University of Iceland's TM dataset, it refers to the WMT-21 dataset. These datasets form separate datasets for English-to-Icelandic and Icelandic-to-English translation tasks, collectively referred to as the IS-EN-NEWS dataset. When training a specific task, we utilized the corresponding training and test sets from the IS-EN-NEWS dataset for both training and evaluation purposes.

For evaluating the fine-tuning results on the Halu-CoT-JSON dataset, we utilized the unused portion of the HaluEval dataset as the test set. The metric employed was the accuracy of JSON adherence, measuring whether the model-generated JSON complies with the task's required schema.

In assessing the fine-tuning results on the IS-EN-NEWS dataset, we employed the test set from WMT-21 to calculate the model's BLEURT scores. The IS-EN-NEWS training dataset comprises two segments: the English-to-Icelandic and Icelandic-to-English translation task training datasets from WMT-21, and Icelandic-English bilingual translations longer than 50 words extracted from the University of Iceland's TM dataset. And test datasets are in accordance with WMT-21. Model performance was evaluated on both tasks separately. Following practices from \citet{Garcia2023-dx} for a fairer assessment, we used BLEURT to evaluate model performance. We followed the recommendations from the publicly available Github page \citep{Pu2021-tj} used the BLEURT-20 checkpoint, and employed a 2-shot learning approach.

\subsection{Emergent Abilities from The Perspective of Leverage Learning}

We observe that this phenomenon, previously described as "emergent abilities", has been analyzed in \citet{Schaeffer2023-ts} and is ascribed to nonlinear or discontinuous metrics. We also ascribed to discontinuousness. However, we posit that the discontinuity lies in the abilities themselves, not the metrics. 

The conventional idea suggests that there is no direct correlation between general data and the performance of LLMs on low-resource tasks. If such a correlation existed, a steady increase in general data should lead to a consistent improvement in LLM performance on these tasks, presumably in a linear relationship with the volume of general data utilized in training. Yet, this view struggles to account for the initial performance improvement and subsequent stabilization observed in LLMs as general data volume increases.

This "S" curve behavior can be elucidated through the lens of Leverage Learning. In Leverage Learning, LLMs acquire capabilities discretely, one quanta at a time. LLMs initially learn a small number of quanta, but these are almost inconsequential for meeting the task's requirements, resulting in slow accuracy growth. As training data continually increases, LLMs gradually learn more quanta, leading to a rapid rise in accuracy once these meet part of the task's demands. Eventually, when LLMs learn enough quanta to fully satisfy the task requirements, the growth in accuracy stabilizes.

\subsection{Evaluation on High-resource Tasks for TELL}

While TELL has shown promising results in low-resource settings, its efficacy in high-resource tasks, such as mathematical problem-solving, warrants examination. Based on the principles of Leverage Learning, we hypothesize that TELL might not perform as well in high-resource tasks as it does in low-resource ones. High-resource tasks likely have both their task-specific and non-specific capabilities already well-learned, leaving less room for improvement compared to low-resource tasks, where both types of capabilities can significantly benefit from further learning. Thus, TELL's effectiveness may not be as pronounced in high-resource scenarios. To test this hypothesis, we conducted experiments using the llama2-7b-chat model on the GSM-8K training set.

\begin{figure}[h]
\begin{center}
\includegraphics[width=5.7cm]{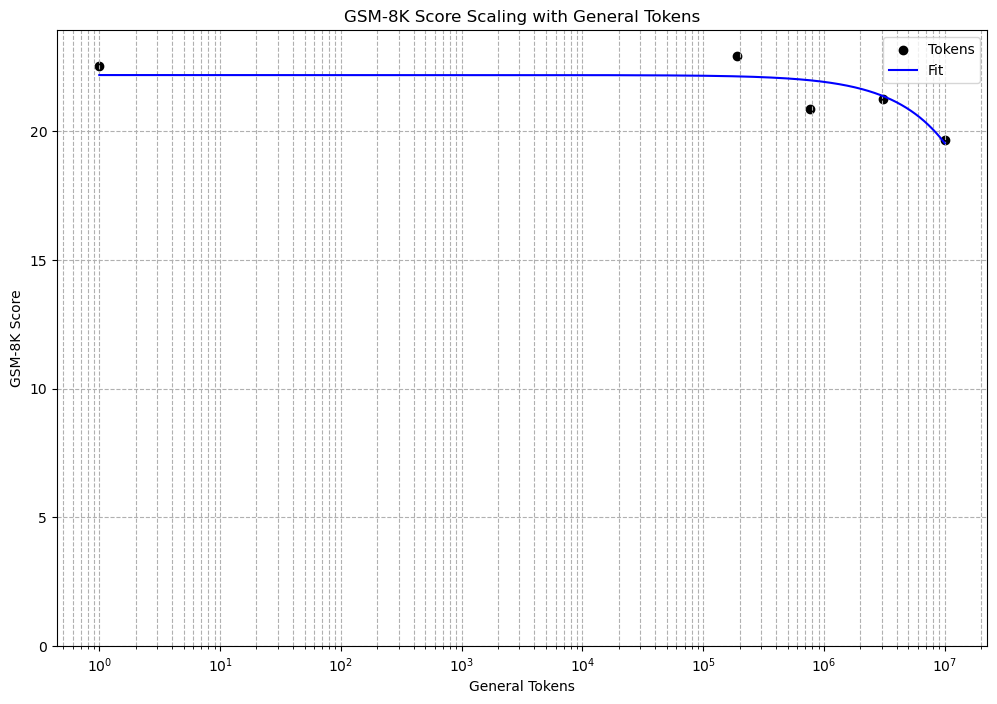}
\end{center}
\caption{GSM-8K score scaling with general tokens}
\end{figure}

The results, validated with the corresponding test set, seem to support our hypothesis.

\subsection{Experimental study on General Dataset for TELL}

In our experiments, we utilized alpaca-en as the general dataset and extended our exploration to other general datasets. The experiments were conducted on the QWen-7b-chat model, where we tested the Open-Platypus dataset released by Platypus2 \citep{Lee2023-yd}. This dataset primarily focuses on code and mathematics, supplemented by a selection of general tasks. We applied the TELL strategy using a comparable number of tokens.

We utilized AGIEval \citep{Zhong2023-um} and OpenBookQA \citep{Mihaylov2018-il} to test general ability. AGIEval \cite{Zhong2023-um} is an exam dataset designed for general human test-takers, including entrance exams for colleges (like China's Gaokao and the US SAT), law school entrance exams, math competitions, and U.S. civil service exams. This project tests the model's ability to world knowledge and understanding of reality. OpenBookQA \cite{Mihaylov2018-il} comprises questions that require multi-step reasoning, common-sense knowledge application, and deep text understanding, testing the model's context comprehension and reasoning capabilities. The arithmetic mean of scores across various sub-categories was used as the final score for this evaluation.

\begin{tabular}{lcc}
\hline
\textbf{Metric} & \textbf{Platypus-en 4e6 tokens} & \textbf{Alpaca-en 4e6 tokens} \\
\hline
agieval-gaokao-chinese & 69.51 & 67.48 \\
agieval-gaokao-english & 89.87 & 85.95 \\
agieval-gaokao-geography & 70.85 & 69.85 \\
agieval-gaokao-history & 85.53 & 81.7 \\
agieval-gaokao-biology & 79.52 & 74.76 \\
agieval-gaokao-chemistry & 47.34 & 53.62 \\
agieval-gaokao-mathqa & 29.34 & 35.04 \\
agieval-logiqa-zh & 50.69 & 39.17 \\
agieval-lsat-ar & 21.74 & 22.17 \\
agieval-lsat-lr & 54.9 & 42.55 \\
agieval-lsat-rc & 56.88 & 54.28 \\
agieval-logiqa-en & 45.47 & 38.71 \\
agieval-sat-math & 40 & 32.73 \\
agieval-sat-en & 80.58 & 72.33 \\
agieval-sat-en-without-passage & 41.75 & 36.89 \\
agieval-aqua-rat & 20.08 & 24.41 \\
agieval-gaokao-physics & 36.5 & 32 \\
agieval-jec-qa-kd & 22.3 & 20.9 \\
agieval-jec-qa-ca & 23.5 & 21 \\
agieval-gaokao-mathcloze & 5.93 & 4.24 \\
agieval-math & 4.3 & 6.8 \\
\textbf{AGIEVAL AVG} & \textbf{46.50} & \textbf{43.65} \\
openbookqa & 72.4 & 67 \\
openbookqa\_fact & 86.6 & 83.4 \\
\textbf{OBQA AVG} & \textbf{79.5} & \textbf{75.2} \\
\textbf{JSON adherence} & \textbf{92.10\%} & \textbf{94.00\%} \\
\hline
\end{tabular}

Our findings indicate that the performance of the TELL strategy on the Open-Platypus dataset was slightly inferior to that on alpaca-en in terms of low-resource task performance. More notably, there were significant differences in general capabilities between the two datasets. Identifying the specific characteristics of general datasets that are most conducive to the TELL strategy remains an area for further investigation.

\subsection{Experimental Study on Self-instruct anchor prompt}

In our study, the anchor prompts were manually designed. However, we believe that the design of anchor prompts can be further optimized and cost-reduced. We experimented with having the LLM used for training automatically generate its anchor prompts. Specifically, we conducted experiments on the IS-EN-NEWS dataset using the Gemma-7b-it model, where Gemma-7b-it generated a prompt to serve as its anchor prompt to guide its learning process.

\begin{figure}[h]
\begin{center}
\includegraphics[width=5.7cm]{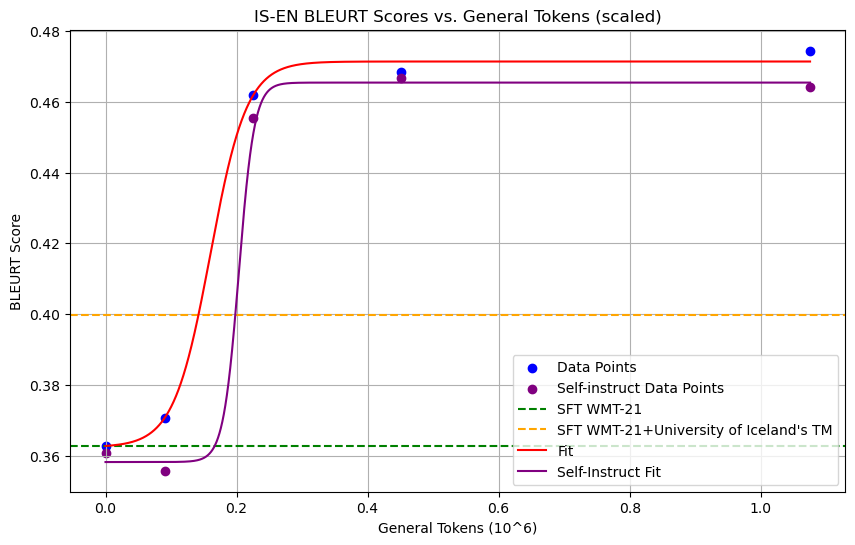}
\end{center}
\caption{Manual prompt vs self-instruct prompt}
\end{figure}

As shown in the graph, the manually designed anchor prompts and LLM self-instructed anchor prompts performed similarly in terms of BLEURT scores. This finding suggests that the design of anchor prompts can be further optimized to reduce costs.

\end{document}